\documentclass{article}

\pdfoutput=1
\PassOptionsToPackage{numbers, compress}{natbib}
\usepackage[numbers]{natbib}

    \usepackage[final]{marneurips_2024}

\usepackage{graphicx}
\usepackage{amsmath}
\usepackage{amssymb}
\usepackage{booktabs}

\usepackage{subcaption}
\usepackage{amsmath}
\usepackage{amssymb}
\usepackage{fontawesome}
\usepackage{enumitem}
\usepackage{adjustbox}
\usepackage{multirow}
\usepackage{colortbl}
\usepackage{arydshln}
\usepackage{xspace}
\usepackage{caption}
\usepackage[most]{tcolorbox}
\tcbuselibrary{skins}
\usepackage{listings}
\usepackage{wrapfig}
\usepackage{array}  
\usepackage{todonotes}
\usepackage{xcolor}

\definecolor{darkgreen}{rgb}{0.0, 0.5, 0.0} 

\usepackage{hyperref}

\definecolor{prompt1}{RGB}{223, 223, 192}
\definecolor{prompt1-frame}{RGB}{137, 137, 90}
\definecolor{prompt2}{RGB}{223, 223, 192}
\definecolor{prompt2-frame}{RGB}{137, 137, 90}
\definecolor{prompt3}{RGB}{212, 238, 179}
\definecolor{prompt3-frame}{RGB}{117, 146, 77}
\definecolor{prompt4}{RGB}{180, 230, 210}   
\definecolor{prompt4-frame}{RGB}{70, 140, 120}  

\tcbset{
    prompt1/.style={
        enhanced,
        breakable,
        colback=prompt1,        
        colframe=prompt1-frame,           
        fontupper=\ttfamily,      
        boxrule=1pt,              
        arc=4mm,                  
        left=1mm, right=1mm, top=1mm, bottom=1mm, 
        boxsep=4pt,               
        before skip=10pt, after skip=10pt, 
        overlay={
            \node[anchor=north west, xshift=4pt, text=white] at (frame.north west) {\faDatabase};
        },
        title={~~~~~~~\textbf{In-context demonstrations for answer parsing.}},
        coltitle=white,
        fonttitle=\bfseries\small
    }
}

\tcbset{
    prompt2/.style={
        enhanced,
        breakable,
        colback=prompt2,        
        colframe=prompt2-frame,            
        fontupper=\ttfamily,               
        boxrule=1pt,                       
        arc=4mm,                           
        left=1mm, right=1mm, top=1mm, bottom=1mm, 
        boxsep=4pt,                        
        before skip=10pt, after skip=10pt, 
        overlay={
            \node[anchor=north west, xshift=4pt, text=white] at (frame.north west) {\faDatabase};
        },
        title={~~~~~~~\textbf{In-context demonstrations for few-shot learning.}},
        coltitle=white,
        fonttitle=\bfseries\small
    }
}

\usepackage[capitalize]{cleveref}
\crefname{section}{Sec.}{Secs.}
\Crefname{section}{Section}{Sections}
\Crefname{table}{Table}{Tables}
\crefname{table}{Tab.}{Tabs.}

\title{Investigating Prompting Techniques for Zero- and Few-Shot Visual Question Answering}

\author{%
  Rabiul Awal \quad Le Zhang\quad Aishwarya Agrawal \\
  Mila - Quebec AI Institute \\ Université de Montréal \\ 
  \texttt{\{rabiul.awal,le.zhang,aishwarya.agrawal\}@mila.quebec}
}

\begin{document}
\maketitle
\begin{abstract}
In this paper, we explore effective prompting techniques to enhance zero- and few-shot Visual Question Answering (VQA) performance in contemporary Vision-Language Models (VLMs). Central to our investigation is the role of question templates in guiding VLMs to generate accurate answers. We identify that specific templates significantly influence VQA outcomes, underscoring the need for strategic template selection. Another pivotal aspect of our study is augmenting VLMs with image captions, providing them with additional visual cues alongside direct image features in VQA tasks. Surprisingly, this augmentation significantly improves the VLMs' performance in many cases, even though VLMs ``see'' the image directly! We explore chain-of-thought (CoT) reasoning and find that while standard CoT reasoning causes drops in performance, advanced methods like self-consistency can help recover it.  Furthermore, we find that text-only few-shot examples enhance VLMs' alignment with the task format, particularly benefiting models prone to verbose zero-shot answers.  Lastly, to mitigate the challenges associated with evaluating free-form open-ended VQA responses using string-matching based VQA metrics, we introduce a straightforward LLM-guided pre-processing technique to adapt the model responses to the expected ground-truth answer distribution. In summary, our research sheds light on the intricacies of prompting strategies in VLMs for VQA, emphasizing the synergistic use of captions, templates, and pre-processing to enhance model efficacy.
\end{abstract}

\section{Introduction}
\label{sec:intro}
Visual Question Answering (VQA) is a challenging task that requires models to comprehend both visual and textual inputs to deliver accurate responses \cite{antol2015vqa}. Recent vision-language models (VLMs) pre-trained on webscale image-text data have made significant advancements towards tackling VQA tasks, including surpassing human performance on the popular VQAv2 dataset when fine-tuned on it \cite{chen2022pali,wang2022image,alayrac2022flamingo}.  A key aspect of these models' functionality in VQA tasks is their potential for prompting by taping on their pre-trained foundational knowledge without any need for task-specific fine-tuning~\cite{eichenberg2021magma, alayrac2022flamingo, manas2022mapl, li2023blip}.  
This process involves utilizing specific textual cues to frame the task, varying from simple task descriptions in zero-shot settings to incorporating examples of image-question-answer triplets in few-shot scenarios. 
\begin{figure}[t]
  \centering
  \begin{adjustbox}{width=0.9\linewidth, center}
    \includegraphics{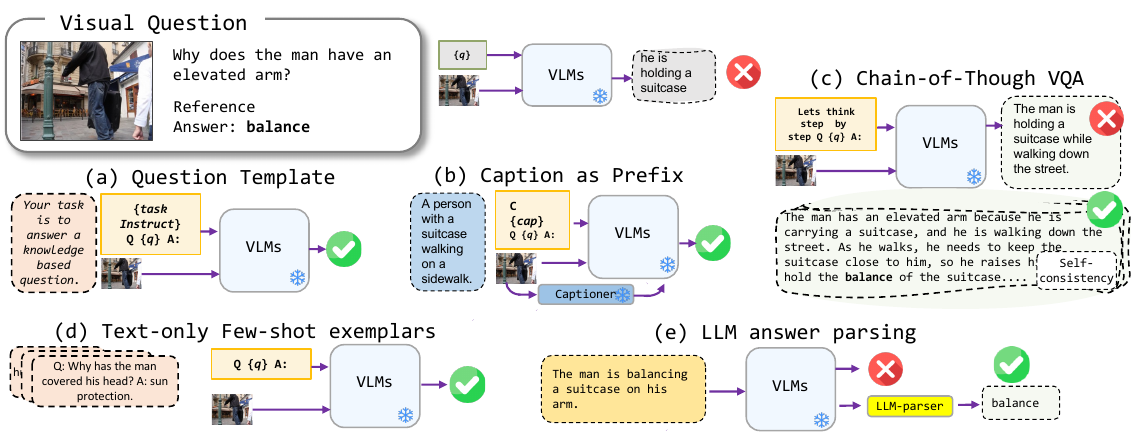}
  \end{adjustbox}

  \caption{Overview of prompting techniques explored with various VLMs, encompassing Standard, Caption, Chain-of-thought VQA and Text-only Few-shot, and the use of LLM-guided pre-processing.}
  \label{fig:overview}
\end{figure}

However, existing works do not systematically evaluate the impact of different prompting techniques on improving the zero-shot/few-shot performance of generative VLMs. As a result, we lack knowledge about which techniques are more effective than others. In this work, we address this gap by conducting a systematic investigation of a wide range of techniques (see Fig.~\ref{fig:overview}), including altering question templates, integrating additional visual cues, implementing chain-of-thought reasoning, and providing text-only few-shot in-context guidance. Additionally, we refine traditional VQA metrics to accommodate these techniques. 

Our goal is to uncover the most effective fine-tuning free prompting techniques for enhancing VQA performance, drawing inspiration from the wide array of prompting methods explored in large language models (LLMs)~\cite{brown2020language, lester2021power, srivastava2022beyond, wei2022chain, kojima2022large}. This exploration is particularly crucial in an era where VLMs, equipped with broad pre-trained knowledge from diverse image-captioning or instruction-tuned datasets, are increasingly being utilized for general-purpose applications. Our investigation includes:

\begin{enumerate}[leftmargin=*]
    \item \textbf{Choice of the Question Template:} We investigate different question templates to guide effective answer generation, by varying the structure and phrasing of questions. Further details are in §\ref{par:technique_1}.

    \item \textbf{Leveraging Captions for Enhanced Context:} We investigate whether zero-shot VQA performance of VLMs can be improved by incorporating image captions as additional visual cues. (§\ref{par:technique_2}).  
    We generate image captions with varying levels of detail and relevance to the question.

    \item \textbf{Incorporating Chain-of-thought Reasoning:}  Inspired by the success of chain-of-thought (CoT) reasoning in language models, we investigate its application in VQA. This approach prompts the model to provide step-by-step rationale alongside answers (§\ref{par:technique_3}). 
     
    \item \textbf{Incorporating Text-only Few-shot Examples:} We incorporate text-only few-shot examples to enhance model performance, particularly in knowledge-based tasks. (§\ref{par:few_shot_exemplars}). 

    \item \textbf{Elevating VQA Metrics for Generative VLMs:}  We improve the traditional string-matching VQA metric with minimal modifications by introducing LLM-based pre-processing to refine verbose model outputs, aligning them with the style of reference answers.
\end{enumerate}

We extensively analyse state-of-the-art VLMs such as BLIP2~\cite{li2023blip}, LLaVa~\cite{liu2023visual}, OpenFlamingo~\cite{awadalla2023openflamingo}, and Kosmos2~\cite{peng2023kosmos}. These open-source models, stemming from diverse training backgrounds like image-conditioned autoregressive pre-training, interleaved image-text pre-training, and instruction-tuning, offer a comprehensive view of current VLM capabilities. We focus on well-established benchmarks like VQAv2~\cite{goyal2017making} and Visual7w~\cite{zhu2016visual7w}, as well as more challenging tasks that involve compositional reasoning (GQA~\cite{hudson2019gqa}) and knowledge-based reasoning (OKVQA~\cite{marino2019ok}, AOKVQA~\cite{Schwenk2022AOKVQAAB}). Additionally, we introduce the recently developed Winoground dataset~\cite{thrush2022winoground} in a VQA format to test models' capabilities beyond the typical COCO distribution.

Our extensive analysis of prompting techniques on four sota VLMs and six VQA benchmarks reveal several key insights: (1) We found that different models have distinct template preferences, indicating that there is not a one-size-fits-all solution. This highlights the importance of careful template selection for optimal performance. (2) 
The incorporation of image captions leads to a noticeable improvement in VQA performance, however the effectiveness varies depending on the model-dataset combination used. (3) While the initial use of CoT reasoning leads to performance drops, approaches like self-consistency~\cite{wang2022self} offer promising avenues for integrating effective rationales. (4) Few-shot exemplars (text-only) effectively improve model alignment with task formats, particularly in the presence of captions and CoT rationales, but their benefits diminish when LLM-based pre-processing is applied. (5) A notable challenge is observed in the Winoground-VQA task, where most VLMs struggle significantly, highlighting the need for advanced model capabilities to handle visio-linguistic compositionality. (6) Our adoption of LLM-guided pre-processing proves crucial for reliable VQA evaluation, correcting inaccuracies in traditional metrics and enabling a more accurate reflection of model capabilities.

Through these insights, our study aims to advance the understanding of how to better utilize large pre-trained VLMs in VQA, particularly in non-fine-tuning scenarios. We hope this work serves as a reference for future research in zero- and few-shot VQA, highlighting innovative approaches to enhance model performance.

\section{Related Work}
\paragraph{VQA tasks and datasets}
Advancements in Visual-Question Answering (VQA) have been largely driven by a variety of benchmark datasets~\cite{goyal2017making, antol2015vqa, zhu2016visual7w, johnson2017clevr, hudson2019gqa, marino2019ok}. One influential example is the VQA v2 ~\cite{antol2015vqa,goyal2017making} dataset, which includes diverse questions about images, requiring a wide range of visual understanding capabilities from models. Specialized datasets such as GQA~\cite{hudson2019gqa} and CLEVR~\cite{johnson2017clevr} target specific visual reasoning aspects: GQA assesses compositional reasoning, while CLEVR focuses on synthetic visual reasoning. The integration of external knowledge with visual understanding is uniquely tested in the OK-VQA~\cite{marino2019ok} and AOKVQA~\cite{Schwenk2022AOKVQAAB} datasets.

While VLMs have made tremendous progress in tackling VQA datasets such as VQAv2, even surpassing human performance on VQAv2 when fine-tuned \cite{alayrac2022flamingo,li2023blip,chen2022pali}, their ability to tackle more complex datasets such as GQA which requires compositional reasoning and OK-VQA, AOKVQA which require knowledge-based reasoning is limited. In this work, we focus our evaluation on these complex datasets, in zero- and few-shot settings. Additionally, we repurpose the recently released Winoground~\cite{thrush2022winoground} benchmark into a VQA format, introducing Winoground-VQA as a novel measure to test compositional reasoning in a more controlled and stringent environment.


\textbf{Prompting in LLMs} The realm of prompting techniques has been a focal point in adapting LLMs for various unseen NLP tasks \cite{brown2020language,lester2021power}. These techniques typically navigate LLMs towards accurate responses, either by employing \textit{in-context} labeled examples \cite{brown2020language} or by crafting precise task instructions \cite{liu2023pre,zhao2021calibrate,liu2021makes,lu2021fantastically,ouyang2022training}. Recent developments have highlighted the effectiveness of specific prompt templates, like ``Let's think step-by-step" \cite{kojima2022large}, in enhancing reasoning and solving complex tasks. This method, known as Chain-of-Thought (CoT) prompting \cite{wei2022chain}, has been particularly successful in larger scale LMs. To facilitate CoT reasoning in smaller LMs, FLAN T5 \cite{chung2022scaling} was introduced, fine-tuning an 11B LM on a combination of natural instructions and CoT data. For a comprehensive understanding of prompting in NLP, readers are directed to survey works such as \cite{liu2023pre} for general prompting techniques and \cite{qiao2022reasoning} for a focus on reasoning. In line with these developments, our study investigates the application of prompting techniques in multimodal VQA tasks.

\textbf{Multimodal Prompting} Prompting is not well explored in multimodal models as large generative VLMs are relatively new. There are a few different lines of work that apply prompting in different ways.
Early models like Flamingo, MAPL and others \cite{tsimpoukelli2021multimodal,alayrac2022flamingo,manas2022mapl} utilize few-shot in-context learning for task adaptation. Flamingo's dependency on interleaved image-text data for pre-training poses data curation challenges, while MAPL's limited training resources result in lower VQA performance compared to state-of-the-art methods. Newer VLMs such as BLIP2~\cite{li2023blip}, LLaVa~\cite{liu2023visual}, MiniGPT4~\cite{zhu2023minigpt} and Kosmos2~\cite{peng2023kosmos} show promising results in zero-shot VQA prompting, largely due to their extensive pre-training.  These models connect vision encoders with large language models (such as LLama2~\cite{touvron2023llama}), aiming for general-purpose visual and language understanding. Notably, LLaVa and MiniGPT4's efforts to emulate GPT-4's multimodal capabilities in dialogue and reasoning mark a significant development, though their effectiveness in zero-shot applications similar to GPT-4 is yet to be fully explored. 

Another emerging approach involves prompting GPT-3~\cite{brown2020language} or Codex~\cite{chen2021evaluating} API in frameworks such as ViperGPT~\cite{suris2023vipergpt} and VisualProg~\cite{gupta2022visual}, which transform complex language queries into executable programs using multiple vision-language models as subroutines. Similarly, approaches such as  PICa~\cite{jin2021good}, PromptCap~\cite{hu2022promptcap} and Img2LLM~\cite{guo2022images} convert images into text descriptions for LLM processing. However, their dependence on GPT-3 API for optimal performance introduces challenges in accessibility and reproducibility, or they face limitations with less capable LLMs. We extend this language-mediated VQA approach to VLMs, where both text and image are considered, as opposed to LLM-only methods.  Additionally, recent works by Zhang \textit{et al.}~\cite{zhang2023multimodal} and  LLaVA~\cite{liu2023visual} employing multimodal CoT reasoning have demonstrated improved accuracy in ScienceQA~\cite{lu2022learn} tasks. However, these successes, primarily due to model fine-tuning on multimodal CoT data, lack extensive evaluation in zero-shot reasoning scenarios. We address these gaps by focusing on fine-tuning-free prompting techniques within accessible VLMs, aiming to uncover effective prompting strategies for various VLM families across a range of tasks.

\textbf{VQA Evaluation} While the original VQA Accuracy metric by Antol \textit{et al.} \cite{antol2015vqa} has been the standard, it faces challenges with generative models due to verbose outputs \cite{agrawal2022rethinking}. A recently proposed method replaces traditional metrics with a fully LLM-based metric capable of handling verbose model outputs without relying on string matching \cite{manas2023improving}. In this work, we introduce a simple LLM-based pre-processing step to make the conventional VQA metric compatible with generative models.


\section{Prompting VLMs for VQA}

\label{par:prompt_settings}
This section provides an overview of various fine-tuning-free prompting techniques aimed at enhancing multimodal zero- and few-shot VQA performance. Drawing inspiration from NLP literature, we adapt these methods to the unique demands of multimodal VQA. Our exploration covers a spectrum of techniques, from altering question templates and integrating additional visual cues to implementing chain-of-thought reasoning and few-shot in-context guidance. The efficacy of these methods is evaluated across diverse open-source VLMs, aiming to bridge knowledge gaps in applying NLP-inspired prompting in multimodal contexts and assess their effectiveness in VQA tasks.
        
        
        


\textbf{Prompting Technique 1: Varying the Question Template}\label{par:technique_1}
This technique involves modifying question templates to guide VLM responses. The aim is to examine how varying the structure and phrasing of questions can affect the model's answer generation process. We refer to this setting as Standard VQA. From a broader range of initial templates, we narrowed down to five key ones: `Null', `follow-qa', `qa', `short-qa', and `instruct-qa'. Each template induces a specific response style; for instance, `Null' and `follow-qa' deviate from the standard ``Question: Answer" format, `short-qa' prompts concise responses, and `instruct-qa' provides task-specific directions.

\begin{table}[]
    \centering
    \begin{tabular}{ll}
        \textbf{Name} &  \textbf{Template} \\
        \midrule
        \multicolumn{2}{l}{\textbf{(1) Standard VQA Templates}} \\
        Null & \{question\} \\
        qa/short-qa & Question: \{q\} \{o\} \textcolor{gray}{[Short]} Answer: \textcolor{gray}{[yes or no?]} \\
        follow-qa & Answer the following question. \{q\} \{o\} \\
        instruct-qa & \{task instruction\} Question: \{q\} \{o\} Answer: \\[0.5em]
        
        \multicolumn{2}{l}{\textbf{(2) CoT VQA Templates}} \\
        reason-qa & Answer the following question by reasoning step-by-step. Q: \{q\} A: \\
        think-qa & Q: \{q\} A: Let's think step-by-step \\[0.5em]
        
        \multicolumn{2}{l}{\textbf{(3) Caption VQA Templates}} \\
        & Context: \{s\} \{apply any VQA template\} \\
        
        \midrule
        \multicolumn{2}{l}{\textbf{Image Captioning Templates}} \\
        a-photo-of & A photo of \\
        q-guided-cap & Describe the image according to the following question \{q\} \\
        \bottomrule
    \end{tabular}
\caption{Instruction templates for VQA and image captioning tasks. Here, \{q\} stands for question, \{o\} represents options, and \{s\} denotes a statement or description related to an image.}
\label{tab:templates}
\end{table}

\textbf{Prompting technique 2: Feed caption as additional input}\label{par:technique_2} 
This technique involves providing image captions as additional input to VLMs. The goal is to assess whether this supplementary textual information can enhance the models' comprehension of the visual content and improve their VQA performance. We refer to this setting as Caption VQA. In our study, we utilize two main captioning templates: `a-photo-of' for initiating captions with a straightforward image description, and `q-guided-cap,' inspired by PromptCAP~\cite{hu2022promptcap}, for generating captions directed by the associated question. Our caption generation employs three distinct models to cover a variety of strategies. The first model, BLIP2~\cite{li2023blip}, is used for LLM-guided dense captioning, sampling multiple captions refined by an LLM for more concise and comprehensive visual descriptions. The second, Kosmos-2~\cite{peng2023kosmos}, focuses on grounded captioning, generating captions that provide precise entity localization within the image. The third strategy employs PromptCap~\cite{hu2022promptcap} for question-guided captioning, ensuring the generated captions are relevant to the query’s subject matter. Further details on caption generation are provided in Appendix \ref{app:cap_gen_methods}.

\textbf{Prompting technique 3: CoT reasoning}\label{par:technique_3} 
In this technique, we prompt VLMs to elicit CoT reasoning, producing both a rationale and an answer. The focus is on examining whether the current capabilities and sizes of VLMs facilitate effective CoT reasoning in complex VQA tasks. We also explore the integration of \textit{self-consistency}~\cite{wang2022self}, an advanced method that generates multiple reasoning paths to potentially improve CoT reasoning. We refer to this setting as CoT VQA. We use two distinct CoT templates from CoT literature. 

\textbf{Prompting technique 4: Providing text-only few-shot examples} \label{par:few_shot_exemplars} 
To further enhance VLM performance, we recognize variations in their ability to handle few-shot examples consisting of image, question, and answer triplets. Notably, OpenFlamingo~\cite{awadalla2023openflamingo} excels in learning from in-context image-text pairs, distinguishing it from other models. However, not all models are capable of utilizing image-text few-shot examples. As reported in the respective papers (and confirmed in our early experiments), except OF, the other VLMs (including BLIP2), do not exhibit any benefits from the incorporation of image-text examples. However, all VLMs can glean context from text-only few-shot examples. Thus, to boost VLM performance, we introduce \emph{text-only} few-shot exemplars. These exemplars provide precise guidance to align the model with the desired task format. For example, when answering questions like `Where are these animals found?' in knowledge-based tasks, specifying details like `Africa' instead of `wild' is crucial for correctness. We select relevant exemplars from the training set for each test question, avoiding overly similar examples to encourage appropriate responses. This strategy can be combined with the techniques in §§\ref{par:technique_1}, \ref{par:technique_2}, and \ref{par:technique_3} to enhance performance across VQA tasks. Full details can be found in Appendix ~\ref{app:few-shot}.

\textbf{Mitigating VQA Metric Challenges Using LLM}
Traditional string-matching VQA metrics face challenges when evaluating VLMs, particularly given the contrast between verbose model outputs and concise VQA reference answers (some failure cases are shown in Appendix \ref{app:vqa_metric_failure_modes}). We identify that the VQA metric can be effectively fixed with minimal modifications. To ensure compatibility with established evaluation practices, we introduce a simple LLM-based pre-processing step. This step involves parsing concise answers, a task that can be successfully accomplished using a publicly available 7B LLM. This approach is more accessible and less complex than deploying a full LLM-based metric~\cite{manas2023improving}, which requires complex reasoning to match lengthy model responses against reference answers. This straightforward LLM-based implementation improves evaluation accuracy and reliability, ensuring that VLM capabilities align with performance metrics while maintaining consistency with traditional evaluation.

\section{Experimental Setup}

\paragraph{Vision-language Models}

Our study undertook an extensive evaluation of various VLMs. The focus was on two variants of the BLIP2~\cite{li2023blip} model, differentiated by their underlying language models: OPT and Flan T5. The BLIP2 models integrated with OPT language models are represented as BO (2.7B) and BO (6.7B), while those paired with the Flan T5 language model in XL and XXL sizes are referred to as BF (XL) and BF (XX), respectively. We also evaluated the OpenFlamingo~\cite{awadalla2023openflamingo} model with 4B parameters in its standard OF form and its instruction-tuned variant, OF(I), to assess the impact of instruction-focused training on VQA performance. The evaluation also included the LLaVa~\cite{liu2023visual} model, featuring the Vicuna (13B) variant, and Kosmos2~\cite{peng2023kosmos}, selected for their distinctive pre-training datasets (visual instruction and grounded image-text) and the less focus on VQA benchmark evaluation.


\paragraph{Datasets}
We evaluate on five VQA datasets and the visio-linguistic probing dataset Winoground, each distinct: (1) VQAv2~\cite{goyal2017making} for real-world image-based Q\&A; (2) Visual7W~\cite{zhu2016visual7w}, focused on object-level Q\&A; (3) OKVQA~\cite{marino2019ok}, emphasizing knowledge-based Q\&A; (4) AOKVQA~\cite{Schwenk2022AOKVQAAB}, requiring commonsense reasoning; (5) GQA~\cite{hudson2019gqa}, evaluating visual compositional reasoning; and (6) Winoground-QA, a novel adaptation of Winoground for visio-linguistic compositional reasoning, repurposed into a yes/no VQA task.  Winoground~\cite{thrush2022winoground} presents two images and two captions for each sample, and the task is to determine the correct image-caption matching, with each caption matching only one image and vice versa. We rephrase the captions as yes/no questions using ChatGPT (see Appendix \ref{app:winoground-qa} for more details). Thus, each sample of Winoground-QA requires answering two yes/no questions for each of the two images. This diversity aims to comprehensively test both in-distribution and out-of-distribution VQA capabilities in MLLMs.

\paragraph{Evaluation Metrics}
In our evaluation, we consider two settings: open-ended and multiple-choice. In the open-ended setting, the VLM is conditioned on the question and the image, while in the multiple-choice setting, it additionally uses provided multiple choices. We evaluate using VQA accuracy \cite{antol2015vqa} for datasets with multiple answers (OKVQA, AOKVQA) and binary accuracy (1/0) for datasets with single answers (multi-choice AOKVQA, GQA). Preprocessing with the Zephyr-7B model includes lemmatization and removing prepositions, articles, and punctuation. For Winoground-QA, we use binary accuracy (1/0) based on four yes/no questions. For Winoground-QA, we use binary accuracy (1/0) based on four yes/no questions. Before performing string matching, we preprocess the generated outputs using the Zephyr-7B model\footnote{\url{https://huggingface.co/HuggingFaceH4/zephyr-7b-alpha}}.

\section{Results}

\begin{figure*}[ht]
  \centering
  \begin{adjustbox}{width=\linewidth, center}
  \includegraphics{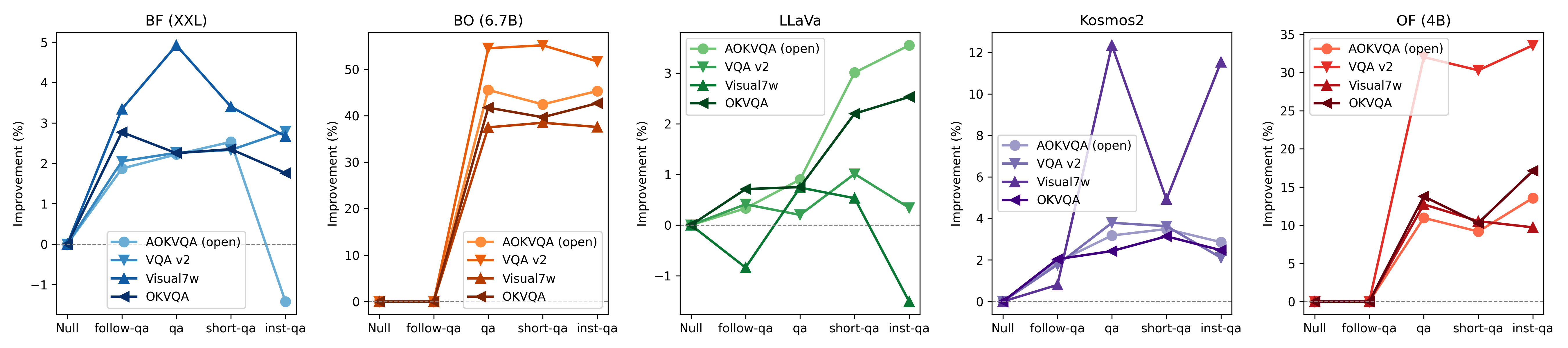}
  \end{adjustbox}
  \caption{\textbf{Comparison of zero-shot VQA} performance across datasets using different templates in the standard setting. All tested models exhibit sensitivity to template variations, as demonstrated by the varying performance improvements over the baseline `Null' template.}
  \label{fig:template_effective}
\end{figure*}
\subsection{Is VQA performance sensitive to the choice of the question template?} \label{sub:result1}

In this analysis of zero-shot VQA performance, presented in Fig.~\ref{fig:template_effective}, we assess the sensitivity of model performance to the choice of question templates. We find a notable variance in performance across different models when applying different templates. For instance, all the models exhibit a significant performance differential of nearly $2$ to $3\%$ between their most and least effective templates, indicating a high sensitivity to question framing. Notably, BO and OF models show a drastic drop to $0\%$ accuracy with non-standard templates, emphasizing the importance of a ``Question: Answer" format in the template used. Conversely, larger BLIP2 models demonstrate reduced sensitivity to template variations, with the Kosmos2 model exhibiting the most significant performance gap of $\sim5\%$ on average.  Interestingly, the optimal template identified as `qa' and `short-qa' for BLIP2 models, there are three cases out of four where the best template is different from author used ones. The variability in model responsiveness to templates underscores the need for tailored approaches, as a one-size-fits-all strategy may not work.  However, the `Null' template consistently underperforms across all models, highlighting the necessity for well-structured prompts in zero-shot VQA. Therefore, our findings suggest that while each model has its unique preferences, employing \textbf{well-optimized templates is crucial for best performance in zero-shot VQA tasks.}

\begin{table*}[ht]
    \scriptsize
    \centering
    \begin{adjustbox}{width=\linewidth}
        
    \begin{tabular}{cp{2cm}cccccccc}
    \toprule
        \bf Benchmark & \bf Strategy & \bf BF (XL) & \bf BF (XXL) & \bf BO (2.7B) & \bf BO (6.7B) & \bf Kosmos2 & \bf LlaVa & \bf OF & \bf OF(I) \\
        \midrule
        \multirow{5}{*}{OKVQA} & standard & 47.43 & 50.13 & 37.73 & 42.7 & 40.33 & 45.77 & 17.11 & 18.29  \\
        & + dense-caption & 46.73 & 48.21 & 37.18 & 43.57 & 40.86 & 44.28 & 33.0 & 34.91\\
        & + grounded-caption & 46.85 & 48.69 & 37.63 & 41.68 & 38.99 & 45.43 & 31.15 & 35.94 \\
        & + PromptCap & \bf 49.07 & \bf 50.55 & \bf 39.81 & \bf 46.29 & \bf 43.09 & \bf 48.01 & \bf 37.38 & 42.48\\
        \cdashline{2-10}
        & \textbf{\textcolor{blue}{\emph{LLM-only}}} \\
        \cdashline{2-10}
        & + dense-capton & 39.04 & 41.87 & 31.14 & 31.14 & - & 35.22 & 29.63 & 36.18\\
        & + grounded-caption & 39.96 & 42.05 & 30.9 & 30.9 & - & 36.0 & 29.7 & 37.36\\
        & + PromptCap & 45.3 & 47.95 & 41.59 & 41.59 & - & 44.74 & 36.37 & \bf 42.62\\

        \midrule
        \multirow{9}{*}{AOKVQA} & standard & 50.68 & 54.66 & 39.89 & 45.57 & 40.85 & 52.69 & 13.57 & 17.27  \\
        & + dense-caption & 49.58 & 51.09 & 37.77 & 45.14 &41.05 & 51.22 & 30.78 & 34.72\\
        & + grounded-caption & 48.50 & 50.53 & 37.59 & 42.52 & 40.11 & 48.20 & 30.00 & 35.15 \\
        & + PromptCap & \bf 52.53 & \bf 55.78 & \bf 43.29 & \bf 49.39 & \bf 43.60 & \bf 52.32 & 39.05 & \bf 44.13 \\
        \cdashline{2-10}
        & \textbf{\textcolor{blue}{\emph{LLM-only}}} \\
        \cdashline{2-10}
        & + dense-caption & 39.11 & 40.23 & 28.78 & 28.78 & - & 35.38 & 30.00 & 35.75 \\
        & + grounded-caption & 37.98 & 39.24 & 25.73 & 25.73 & & 32.49 & 28.01 & 35.42\\
        & + PromptCap & 46.98 & 49.39 & 42.51 & 42.51 & - & 45.75 & \bf 39.27 & 43.96\\
        \midrule
        
        \multirow{3}{*}{GQA} & standard & 44.56 & 45.25 & 35.83 & 38.46 & 37.33 & 38.40 & 28.44 & 26.37 \\
        & + dense-caption & 42.45 & 42.78 & 35.71 & 37.01 & 36.75 & 36.03 & 33.44 & 33.16 \\
        & + grounded-caption & 44.08 & 43.79 & 36.93 & 37.13 & 35.71 & 39.34 & 34.16 & 33.75 \\
        & + PromptCap & \bf 46.60 & \bf 47.01 & \bf 39.08 & \bf 40.32 & \bf 40.13 & 41.00 & \bf 38.04 & \bf 40.00 \\
        \cdashline{2-10}
        & \textbf{\textcolor{blue}{\emph{LLM-only}}} \\
        \cdashline{2-10}
        & + dense-caption & 40.69 & 40.51 & 26.25 & 32.33 & - & 32.33 & 29.89 & 33.12 \\
        & + grounded-caption & 40.22 & 39.21 & 24.28 & 32.76 & - & 32.76 & 29.72 & 33.75 \\
        & + PromptCap & 45.70 & 45.68 & 36.46 & 40.34 & - & \bf 43.01 & 34.95 & 38.89 \\
        \midrule
        \multirow{3}{*}{VQAv2} & standard & 64.22 & 66.66 & 54.1 & 54.53 & 53.52 & 56.2 & 33.58 & 35.41 \\
        & + dense-caption & 63.1 & 65.25 & 54.58 & 55.78 & 47.26 & 59.18 & 45.35 & 47.98 \\
        & + grounded-caption & 63.13 & 65.16 & 52.56 & 55.33 & 45.63 & 56.81 & 44.94 & 45.24 \\
        & + PromptCap & \bf 70.7 & \bf 71.37 & \bf 58.78 & \bf 62.81 & \bf 57.33 & 65.32 & \bf 56.93 & 58.0 \\
        \cdashline{2-10}
        & \textbf{\textcolor{blue}{\emph{LLM-only}}} \\
        \cdashline{2-10}
        & + dense-caption & 55.51 & 57.15 & 39.68 & 39.68 & - & 50.22 & 43.47 & 49.42\\
        & + grounded-caption & 55.02 & 56.49 & 34.6 & 34.6 & - & 48.97 & 41.89 & 49.6 \\
        & + PromptCap & 69.14 & 68.3 & 57.61 & 57.61 & - & \bf 66.58 & 55.1 & \bf 60.26 \\
    \bottomrule
    \end{tabular}
    \end{adjustbox}
    \captionsetup{font=small}
    \caption{Caption VQA performance across VLMs with additional visual contexts: dense captioning, visual grounding, and question-aware captioning (PromptCap). Bold values indicate the best performance, highlighting the benefits of added visual context.}
    \label{tab:caption_vqa}
\end{table*}
\subsection{Augmenting VLM's context with image captions and LLM-only VQA results}

We investigate how different qualities and types of image captions, presented as text-based visual cues, impact zero-shot VQA performance in VLMs. Our evaluations address six specific questions (Q1-4  in Table~\ref{tab:caption_vqa}, along with Q5 and Q6 described in Tables~\ref{tab:ablation_image_feats} and \ref{tab:winoground_qa}, respectively).

\paragraph{Q1. Can VLM effectively utilize image captions in-context with its language model alone?}

Answer: \textbf{Yes.} The results in Table \ref{tab:caption_vqa} indicate that VLMs can effectively leverage quality in-context information, leveraging the strengths of language modality alongside patch-level features. However, the degree of \textbf{improvement varies depending on the model-dataset combination and task difficulty}. For instance, on OKVQA, the best captioning technique PromptCap enhances the BO (6.7B) model's accuracy by $4.54\%$, while on AOKVQA, the improvement is slightly lower at $3.82\%$. Notably, dense and grounded captioning methods exhibit variability in effectiveness. While less performant models benefit significantly from generic captions, stronger models like BLIP2 can be negatively impacted by low-quality in-context information. This variability suggests that in-context information needs to amplify the inherent image features of VLMs as effectively as specialized methods like PromptCap. Furthermore, our analysis indicates varying levels of improvement across different benchmarks, with significant gains observed in tasks like VQAv2, but less pronounced benefits in tasks requiring multi-step inference or compositional reasoning, such as GQA.

\textbf{Q2. Does plugging an instruction-tuned LLM versus non-instruction-tuned with the same vision backbone matter for in-context learning?} \emph{Test: instruction-tuned model (e.g., OF (I)) vs non-instruction-tuned model (e.g., OF).} Answer: \textbf{Yes}, instruction-tuned models consistently outperform their non-instruction-tuned counterparts across all captioning methods, even though both uses the same vision backbone. This highlights the clear advantage of choosing an instruction-tuned LLM to create a multimodal model. Additionally, the state-of-the-art performance of the BF (XXL) model (instruction-tuned Flan T5 LLM) across various datasets further emphasizes the strength of instruction-tuned models.

\textbf{Q3. Does caption VQA outperform standard VQA across question types?} \emph{Test: Caption VQA gains across question types vs the standard VQA (=baseline)} Answer: \textbf{Yes.} Caption VQA consistently outperforms the standard VQA baseline across various question types in the VQAv2 benchmark with the BF (XXL) model. Notable improvements are observed in questions involving numerical values ($39\%$), color recognition ($20\%$), counting ($13\%$), brand identification ($11\%$), and object identification ($6\%$). However, limited improvements were seen in binary e.g. ``yes/no''  ($-4\%$), complex reasoning e.g. ``why'' ($-1\%$), and localization e.g. ``where'' ($0.96\%$) questions. This discrepancy suggests the potential for integrating additional, targeted techniques specifically designed to handle questions requiring abstract reasoning and spatial understanding.

\textbf{Q4. Can LLM-only models with visual cues in text suffice for VQA compared to VLMs using the same LLM?} \emph{Test: LLM-only model with no access to direct image features vs VLMs augmented with image captions.} Answer: \textbf{Not really.} While LLM-only models with visual cues show promising performance, they are outperformed by VLMs. For instance, on the GQA benchmark, VLMs with PromptCap enhancement achieve up to $40.32\%$ accuracy, significantly higher than LLM-only counterparts. However, within the LLM-only setup, all captioning techniques improve performance, with the quality of in-context information directly correlating with gains. No surprise PromptCap emerges as the most effective, achieving $36.46\%$ accuracy in GQA. Interestingly, dense and grounded captioning also show comparable gains in the LLM-only setup, indicating their utility as a proxy, particularly when direct image features are absent.

\begin{table}[h]
    \centering
    \begin{tabular}{lcccc}
        \toprule
        & \bf BF (XXL) & \bf BF (6.7B) & \bf Kosmos2 & \bf LLaVa  \\
        \midrule
        image & 51.09 & 49.39 & 43.60 & 51.22 \\
        zeroed-image & \textcolor{red}{38.37} & 44.83 & 41.84 & \textcolor{red}{37.96} \\
        \bottomrule
    \end{tabular}
    \caption{\textbf{Effect of nullifying input image} on VLMs in AOKVQA.}
    \label{tab:ablation_image_feats}
\end{table}

\textbf{Q5. Are VLMs using patch-level features with the presence of captions?} \emph{Test:  remove image features from VLMs while retaining captions vs keep both.} Answer: \textbf{Yes, definitely}. Table ~\ref{tab:ablation_image_feats} results show that image features are indispensable, as performance significantly decreases when they are omitted. For instance, the LLaVa model's accuracy drops significantly from $51.22\%$ with image features to $37.96\%$ without, underscoring the critical role of patch-level image features.

In summary, our analysis underscores the beneficial role of captioning techniques in enhancing VLMs for zero-shot VQA, with PromptCap leading the way. It also highlights the value of dense and grounded captioning, especially in LLM-only contexts. However, there is variability in performance across dataset-model combinations when integrating additional visual cues to optimize VQA performance.

\begin{table*}[h]
    \begin{minipage}{0.49\linewidth}
        \centering
        \begin{adjustbox}{width=\linewidth}
        \begin{tabular}{lcccc}
            \toprule
            Model & OKVQA & AOKVQA & GQA & VQA v2  \\
            \midrule
            BF (XL) & 38.98 & 45.96 & 36.56 & 49.94 \\
            BF (XXL) & \bf 42.12 & \bf 47.40 & \bf 39.32 & \bf 55.65 \\
            LLaVa & 33.22 & 45.61 & 30.50 & 47.97 \\
            \bottomrule
        \end{tabular}
        \end{adjustbox}
        \captionsetup{font=small}
        \caption{Results of CoT VQA (Q $\rightarrow$ RA) on open-ended VQA answers. We report the best results across the two CoT templates.}
        \label{tab:cot_qa}
    \end{minipage}
    \hfill
    \begin{minipage}{0.49\linewidth}
        \centering
        \begin{adjustbox}{width=\linewidth}
        \begin{tabular}{llc}
            \toprule
            Method & Format & Accuracy \\
            \midrule
            CoT & Q $\rightarrow$ RA & 47.40 \\
            CoT-iterative & QR $\rightarrow$ A  & 44.93  \\
            CoT-context & RQ $\rightarrow$ A & 49.94  \\
            CoT-consistency $(t=0.7)$ & VOTE(QRi $\rightarrow$ Ai) & \bf 54.53 \\
            \bottomrule
        \end{tabular}
        \end{adjustbox}
        \captionsetup{font=small}
        \caption{Self-consistency CoT narrows the performance gap with standard VQA on AOKVQA when using the BF (XXL) model.}
        \label{tab:cot_improve}
    \end{minipage}
    \\
    \begin{minipage}{0.49\linewidth}
        \centering
        \begin{adjustbox}{width=\linewidth}
        \begin{tabular}{lcccc|c}
            \toprule
            & BF (6.7B) & BO (6.7B) & LLaVa & Kosmos2 & \textcolor{green}{OF} \\
            \midrule
            \multirow{2}{*}{Standard VQA} & \textcolor{gray}{-0.12} & \textcolor{gray}{0.26} & \textcolor{gray}{7.87} & \textcolor{gray}{16.34} & \textcolor{gray}{11.01}\\ 
             & -2.86 & -6.8 & -4.84 & 0.47 & 5.25\\ 
            \cline{2-6}
            \multirow{2}{*}{Caption VQA} & \textcolor{gray}{0.79} & \textcolor{gray}{6.7} & \textcolor{gray}{17.72} & \textcolor{gray}{20.63} & \textcolor{gray}{24.05} \\ 
             & -1.48 & 0.74 & -0.67 & -1.31 & 4.98\\ 
            \cline{2-6}
            \multirow{2}{*}{CoT VQA} & \textcolor{gray}{2.35} & \textcolor{gray}{-} & \textcolor{gray}{5.39} & \textcolor{gray}{-} & \textcolor{gray}{-} \\
             & 2.52 & - & 2.28 & - & - \\
            \bottomrule
        \end{tabular}
        \end{adjustbox}
        \captionsetup{font=small}
        \caption{Few-shot vs. Zero-shot performance on AOKVQA, with (\textcolor{gray}{highlighted}) and without LLM pre-processing.}
        \label{tab:few_shot_vqa}
    \end{minipage}
    \hfill
    \begin{minipage}{0.49\linewidth}
        \centering
        \begin{adjustbox}{width=\linewidth}
            \begin{tabular}{p{1.75cm}ccccc}
                \toprule
                Model & standard & dense & grounded & promptcap  \\
                \midrule
                BF (XXL) & \bf 7.75 & \bf 9.25 & \bf 9.00 & \bf 7.75 \\
                BO (6.7) & 0.0 & 0.0 & 0.0 & 1.0 \\
                LLaVa & 1.25 & 2.0 & 2.5 & 2.25 \\
                Kosmos2 & 0.75 & 0.75 & 0.0 & 1.25 \\
                OF & 0.0 & 0.25 & 0.75 & 0\\
                OF (I) & 0.0 & 0.75 & 0.0 & 0.25 \\
                \midrule
                Random chance & \multicolumn{4}{c}{6.25} \\
                \bottomrule
            \end{tabular}
        \end{adjustbox}
        \caption{\textbf{Performance on the Winoground-VQA} task for both the Standard VQA and Caption VQA settings.}
        \label{tab:winoground_qa}
        \end{minipage}
    
\end{table*}

\textbf{Q6. How do models perform on the Winoground-VQA task?} Answer: \textbf{Very Poorly.} Table~\ref{tab:winoground_qa} contain results on Winoground-VQA, containing both Standard VQA and Caption VQA. Our findings strikingly mirror the observations made in the original Winoground study. The majority of tested models struggle significantly with this task, achieving near-zero accuracy, even the visual instruction-tuned LLaVa that boasts complex reasoning capabilities. Interestingly, for BLIP2, incorporating Caption VQA results in a slight performance improvement. A common trend among many models is their inclination to default to a `yes' response for most questions. This trend may stem from the models grappling with the out-of-distribution characteristics of the Winoground-VQA questions and a potential language bias favoring `yes' answers. This finding is particularly significant considering these models' otherwise strong performance in established VQA benchmarks.


In summary, although our results mainly concentrate on image captions, they highlight the wider scope of utilizing other types of in-context information sources to overcome VLM limitations and fulfil VQA task demands. For instance, integrating an object-detector-based counting engine can be helpful for \textit{counting} questions.  However, it's essential to consider the specific challenges posed by out-of-distributional compositionality when applying these strategies.

\subsection{Do VLMs showcase CoT reasoning in VQA?}


We investigate zero-shot CoT rationales for VQA accuracy in VLMs trained on instruction-tuning datasets. Our experiments focus on LLaVa and BF models, which generate zero-shot rationales (Table 3). Surprisingly, despite sharing a 13B LLM base, BF outperforms LLaVa, but both underperform compared to standard VQA. Qualitative analysis reveals LLaVa's \emph{lengthy inconsistent rationales and hallucinations}, highlighting challenges in robust multimodal reasoning. Our findings question the complex reasoning capabilities of VLMs (as contended in recent models ~\cite{liu2023visual,zhu2023minigpt}), contrary to successes in LLM-only CoT \textbf{at the tested model scale}. Further analysis in Appendix ~\ref{app:cot_samples} contains qualitative examples for CoT.

To improve the effectiveness of rationalization further, we explored three key modifications in \autoref{tab:cot_improve}:
a) \textbf{CoT-iterative}, where we trim reasoning chains to one sentence and condition the final answer on this concise rationale, addressing issues of hallucinations in longer chains. b) \textbf{CoT-context}, which entailed reordering input by placing the generated rationale before the question, slightly improving performance; and c) \textbf{CoT-consistency}, inspired by Wang \textit{et al.'s}~\cite{wang2022self}  self-consistency approach, we sample 30 reasoning paths and adopt a majority vote for the final answer. This method proves most successful, matching performance to the standard VQA setting. In conclusion, while the self-consistency technique is derived from the LLM literature, it shows transfer potential for enhancing the reasoning capabilities of VLMs.

\subsection{Do text-only few-shot exemplars help?}

Table~\ref{tab:few_shot_vqa} shows that text-only few-shot exemplars improve model alignment with the task format. Models like LLaVa and Kosmos2 benefit the most as they tend to generate verbose answers in zero-shot scenarios. This improvement is particularly pronounced in caption and chain-of-thought (CoT) settings, where we provide additional context alongside exemplar questions, helping the models understand the task better and avoid confusion with test questions.

Conversely, the BLIP2 model, which already produces concise answers in zero-shot, does not show substantial improvements with few-shot exemplars. Notably, for OF, we employ image-text few-shot examples (unlike others), and this model consistently performs better across all prompting settings due to its capability to utilize image-text examples.

\subsection{Does LLM pre-processing mitigate the challenges associated with the VQA metric?} 
\begin{table}[h]
    \centering
    \begin{tabular}{lcccc}
        \toprule
        & \bf BF(XXL) & \bf BO(6.7B) & \bf Kosmos2 & \bf LLaVa  \\
        \midrule
       VQA-metric & 50.28$_{3.07}$  & 38.24$_{0.93}$ & \textcolor{red}{13.73$_{74.08}$} & \textcolor{red}{0.38$_{0.09}$} \\
       + LLM Parsing & 53.17$_{2.28}$ & 44.46$_{2.00}$ & 39.67$_{1.42}$ & 50.71$_{2.09}$ \\
       \bottomrule
    \end{tabular}
    \caption{\textbf{LLM-based parsing} stabilizes (red indicates significant failure) VQA accuracy metric across different prompt templates on the AOKVQA dataset.}
    \label{tab:ablation_llm_parsing}
\end{table}

Table~\ref{tab:ablation_llm_parsing} demonstrates the positive outcome of using LLM-guided pre-processing to more accurately reflect the performance of VQA models. Traditional metrics, initially used, fell short in capturing the true capabilities of these models. By implementing a few lines of code for LLM-guided pre-processing prior to applying VQA metrics, we were able to correct the accuracy values for all tested models, leading to a more trustworthy evaluation. This correction proves especially vital for models like LLaVa, which initially displayed unusable metrics. The recalibration also brings a necessary correction to the data for OPT models, addressing misrepresentations in previous reports\footnote{Li \textit{et al.} \cite{li2023blip} reported lower performance figures for OPT variants due to the limitations of traditional VQA metrics.}. Furhtermore, upon closer examination in Table~\ref{tab:few_shot_vqa}, we observe that when LLM-based pre-processing is applied, the performance gain diminishes for all models except OF. The initial improvement can be attributed to the VQA metric's struggle with matching reference answers with generated responses. Few-shot exemplars encourage concise answers, bringing gains during evaluation. Notably, text-only exemplars mainly guide answer format, achievable through pre-processing.

\section{Conclusion}
In summary, our research explores fine-tuning-free prompting strategies to enhance VQA performance for VLMs. We've highlighted the impact of question templates, the benefits of caption prefixes, and the effectiveness of few-shot examples in specific scenarios. Chain-of-thought reasoning had mixed results, but self-consistency helped bridge the gap. Our study provides practical techniques to leverage large pre-trained VLMs for VQA without fine-tuning, contributing to the advancement of zero- and few-shot VQA.

{
    \bibliographystyle{plainnat}
    \bibliography{bib}
}
\appendix

\clearpage
\setcounter{page}{1}

\section{Dataset: Winoground-VQA}
\label{app:winoground-qa}
We have adapted the Winoground dataset for a binary Q\&A task, employing ChatGPT (\texttt{gpt-3.5-turbo})\footnote{\url{https://platform.openai.com/}} for this purpose. To transform a given text into a binary question suitable for the Visual Question Answering (VQA) task, we utilize the following prompt: ``Convert this text into a yes/no question for the Visual Question Answering task:  $<$text$>$.''
To ensure the quality of the generated questions, we have implemented a manual verification process. Questions not meeting the specified quality standards are subject to regeneration. For evaluation, we employ two distinct methods. Firstly, we use prompts for the VQA task that makes use of the text found in the Winoground dataset. For example: ``Does this describe the image? \emph{The taller person hugs the shorter person.}" This approach allows us to evaluate how well the model understands and responds to questions related to the given text. Secondly, we utilize the questions converted through the aforementioned method. For instance: ``Answer the following yes/no question. \emph{Does the taller person hug the shorter person?}"

\begin{table}[h]
\small
\begin{tabular}{p{0.45\linewidth}p{0.45\linewidth}}
\textbf{Original Statement} & \textbf{Converted Question} \\
\midrule
The taller person hugs the shorter person & Does the taller person hug the shorter person? \\
A tree smashed into a car & Did a tree smash into a car? \\
The person without earrings pays the person with earrings & Does the person without earrings pay the person with earrings? \\
The image shows a computer on top of books & Does the image show a computer on top of books? \\
A brown dog is on a white couch & Is a brown dog on a white couch? \\
The happy person is on the right and the sad person is on the left & Is the happy person on the right and the sad person on the left? \\
The heavy oncoming traffic is contrasted with the light outgoing traffic & Is the heavy oncoming traffic contrasted with the light outgoing traffic? \\
A metal chess piece rests on wood objects & Is there a metal chess piece resting on wood objects? \\
Rectangular bushes are behind pointy bushes & Are rectangular bushes behind pointy bushes? \\
\bottomrule
\end{tabular}
\caption{Winoground-VQA. Conversion of original statements to binary questions}
\label{table:winoground_qa}
\end{table}

\section{Experimental Settings}

\subsection{Model Description}

\begin{itemize}[leftmargin=*]
    \item \textbf{BLIP2}~\cite{li2023blip}: BLIP2 combines frozen pre-trained image encoders and large language models with a lightweight, 12-layer Transformer encoder, known as Q-Former, as the only trainable part. It bridges the gap between vision and language models and excels in tasks like image-captioning, leveraging an efficient pre-training strategy that outperforms larger models like Flamingo in zero-shot VQAv2.
    
    \item \textbf{Kosmos2}~\cite{li2023blip}: A Transformer-based causal language model trained on a web-scale dataset of grounded image-text pairs (GRIT). Kosmos-2 excels in multimodal grounding, reducing common language model hallucinations, and is adept at a wide range of tasks, including multimodal referring, perception-language tasks, and language understanding and generation.
    
    \item \textbf{LLaVa}~\cite{liu2023visual}: A large multimodal model combining a vision encoder and Vicuna LLM. LLaVa mimics the capabilities of multimodal GPT-4 through visual instruction tuning and achieves state-of-the-art accuracy on Science QA. It features multimodal chat abilities, including discussing images, identifying objects, and detecting manipulated images.
    
    \item \textbf{OpenFlamingo~\cite{awadalla2023openflamingo}}: An open-source replication of DeepMind's Flamingo models, OpenFlamingo processes interleaved sequences of images and text. It is capable of tasks like captioning, visual question answering, and image classification, achieving similar to Flamingo's performance on various vision-language datasets. The model uses a CLIP ViT-L/14 vision encoder and variants of language models including MPT-7B, outfitting the layers of a pretrained, frozen language model for cross-attention to visual features.
\end{itemize}

\subsection{Caption Generation Strategies (with examples)}\label{app:cap_gen_methods}

We use three specialized models, each chosen for its ability to generate captions that uniquely enhance visual comprehension through text.
\begin{enumerate}[leftmargin=*]
\item \textbf{LLM-guided dense caption with BLIP2:} Leveraging BLIP2~\cite{li2023blip}, we generate multiple captions per image to capture a comprehensive visual description. These captions are then refined into a concise and comprehensive description using the Zephyr 7B LLM, guided by tailored in-context demonstrations.  Example: ``A photo of a room with a green table and chairs. The room also features a green and white kitchen." 

\item \textbf{Grounded caption with Kosmos-2:} The Kosmos-2 model~\cite{peng2023kosmos} generates captions that include specific entities and their locations within the image, leading to more grounded and precise visual descriptions. Example: ``A photo of a kitchen in a dollhouse, with a white stove, sink, and green cabinets (a white stove, sink, green cabinets)."

\item \textbf{Question-guided caption with PromptCap}: This approach leverages the PromptCap~\cite{hu2022promptcap} model, which uses the question to guide caption generation and ensure that the captions closely align with the subject matter of the question. It outperforms generic captions and achieves state-of-the-art accuracy on knowledge-based VQA tasks. PromptCap customizes the caption according to the input question prompt, making it suitable for working with black-box language models like GPT-3 or ChatGPT. Example: ``A photo of a kitchen in a dollhouse.''
\end{enumerate}

\subsection{Few-shot exemplars selection}
\label{app:few-shot}
We have devised a strategy based on the nearest neighbor threshold for selecting five exemplars from the training set for few-shot learning. This approach utilizes the Sentence-BERT (SBERT)\footnote{\url{https://www.sbert.net/}} sentence embedding model to generate embeddings for the questions. Subsequently, we employ cosine similarity to pinpoint the top-k samples that bear the closest resemblance to a specific query. An integral part of our method is the application of a similarity threshold, set at 0.6, to circumvent the selection of samples excessively similar to the query. We've observed empirically that high similarity can inadvertently cause a decline in the model's performance, as the model tends to replicate from the in-context Q\&A pairs instead of generating unique responses to the test query.

Moreover, the content of the \textit{in-context} exemplars varies depending on the specific type of QA task. For Standard VQA, we pair the selected question with its corresponding answer. In the case of Caption VQA, the question is paired with the model-generated caption and its associated answer. For the CoT VQA task, we pair the question with the corresponding model-generated rationale and answer.

\subsection{Samples Illustrating VQA Metric Failure Modes}\label{app:vqa_metric_failure_modes}
\begin{table}[h]
    \centering
    \begin{tabular}{p{9.5cm}p{3cm}}
        \toprule
        \bf Verbose outputs & \bf Reference Answer \\
        \midrule
        The white substance is icing. & icing \\
        A cell phone. & phone \\
        They are surfing on a wave. & surfing \\
        A motorcycle can be used for racing. Racing is a sport. The final answer: racing. & racing \\
        Rainbow cake. The image shows a table with a rainbow cake on it. & rainbow \\
        \bottomrule
    \end{tabular}
    \captionsetup{font=small}
    \caption{Instances where the verbose output of the VQA system, though correct, can not be directly matched with the ground truth using string matching. This discrepancy can lead to misinterpretation of the VQA system's accuracy.}
    \label{tab:vqa_metric_failures}
\end{table}
Table~\ref{tab:vqa_metric_failures} demonstrates examples where discrepancies arise between the verbose outputs of the generative VQA model and the ground truth reference answers. These examples are crucial in highlighting the limitations of conventional VQA metrics. Our approach involves an LLM-guided method capable of parsing verbose answers into a format that aligns with the reference style, thereby accurately evaluating the system's performance.

\subsection{Inference details} For answer generation, we use beam search with a beam size of 3. The maximum token limit is set to 10 for BLIP2 and 50 for verbose models like Kosmos2, with a length penalty of -1 to encourage brevity. For captions and rationales, which require more detail, the length penalty is adjusted to 1 and the maximum token count is increased to 128, balancing informativeness and conciseness.

\section{In-context Demonstrations}

\subsection{Example prompt template: LLM-guided answer parsing}
In Box~\ref{box:in_context_llm_parsing}, we show a sample of the LLM-guided answer parsing template. We use samples in context to guide the language model to produce a short answer e.g. ``two to three words''.

\begin{tcolorbox}[prompt1]
The task is to parse the short answer from input question and long answer. 
The answer should be a max one to three words or a short phrase.\\
Input: What sport can you use this for? 
You can use this motorcycle for off-road sports, such as motocross, enduro, or trail riding. 
Short answer: motocross\\
Input: What area of a school might this be? 
This area of the school might be a library or a classroom, as there are books and chairs in the background. 
Short answer: library\\
Input: What type of bread is this meal made from? 
This meal is made from pita bread. Short answer: pita\\
Input: Which brand of car is shown in this picture? 
The brand of car shown in the picture is a Volkswagen. 
Short answer: Volkswagen\\
Input: Is this a private or public room? This is a public room. 
Short answer: public\\
Input: What is the name of the device that is protecting people from the rain in this picture? 
The device that is protecting people from the rain in this picture is an umbrella. 
Short answer: umbrella\\
Input: Why might someone go to this place? 
Someone might go to this place, which appears to be a busy street in a city, 
for various reasons such as shopping, dining, socializing, or attending events. 
Short answer: shopping\\
Input: How tall do these animals get? Giraffes can grow up to 18 feet tall. 
Short answer: 18 feet\\
Input: What is this desk used for? 
The desk is used for working on a computer, making phone calls, and organizing office supplies. 
The final answer is working. 
Short answer: working\\
Input: How long does this animal usually live? The image shows shep. The average lifespan of a sheep is 10 years. 
Short answer: 10\\
\label{box:in_context_llm_parsing}
\end{tcolorbox}

\subsection{Example prompt template: Few-shot exemplars}
Box~\ref{box:in_context_few_shot} shows a full prompt we utilized to prompt VLM under the Caption VQA setting for the multiple-choice AOKVQA dataset. The task is designed to generate answers pertaining to a specific image. Incorporated within the template are a set of caption-question-answer triplets that are unrelated to the candidate question. These caption-question-answer triplets serve as the context for guiding the model's response. The concluding task for a VLM, guided by the prior examples within the template, is to deliver a knowledgeable and contextually accurate answer to a visual question derived from a specific image.
\begin{tcolorbox}[prompt2]
In this task, your goal is to write an answer to a given question about the image.
To write the answer, here are some sample QA suggestions (not relevant to the image):\\
Context: A photo of a person taking a tray of chocolate muffins out of the oven. 
Question: What is the likely flavor of these muffins? Blueberry, pumpkin, banana or red velvet? 
Answer: Red velvet

Context: A photo of a laptop and a donut on a table the orange mug to the left of the donut is made of plastic. 
Question: What material is the orange mug to the left of the donut made out of? Ceramic, glass, metal or plastic? 
Answer: Glass

Context: A photo of a box of red velvet cupcakes. 
Question: Which cupcake is alcohol-free? Red velvet, cherry amaretto, strawberry daiquiri or bailey's chocolate? 
Answer: Red velvet 

Context: A photo of a little girl eating a piece of cake with white icing. 
Question: The white part of the icing here is likely flavored with what? Onion, vanilla, potato or peppermint? 
Answer: Vanilla 

Context: A photo of a table with plates of breakfast food with yellow fruits on top of the pancake. 
Question: What color are the fruits sliced out on top of the pancake? Red, white, blue or pink? 
Answer: White\\

Now answer the following question about the image. Your task is to answer a knowledge based question.

Context: A photo of a person holding a cupcake with whipped cream on top.  
Question: What is the white substance on top of the cupcakes? Mayo, ice cream, butter or icing?
Answer:
\label{box:in_context_few_shot}
\end{tcolorbox}

\section{Additional Results}
\subsection{Analysis on Quality of Generated CoT rationales}\label{app:cot_samples}
Our analysis of the AOKVQA dataset, detailed in Table~\ref{tab:cot_quality}, sheds light on the performance of the BF (XXL) and LLaVA models in generating explanations compared to human-authored ground truths. The LLaVA model, in particular, is prone to producing longer rationales, potentially influenced by its training on detailed narrative datasets. Several types of errors were noted:
\begin{table}[ht]
    \centering
    \begin{adjustbox}{width=0.5\linewidth}
    \begin{tabular}{llccc}
    \toprule
        Model & Prompt strategy & Rouge-1 & Rouge-L & BERTScore \\
        \midrule
       \multirow{2}{*}{BF (XL)} & CoT & 28.04 & 24.75 & 88.16   \\
        & CoT $(n=5)$ & 26.95 & 24.04 & 88.24 \\
        \hline
        \multirow{2}{*}{BF (XXL)} & CoT & 29.54 & 26.68  & 88.05  \\
        & CoT $(n=5)$ &  28.08 & 25.51 & 87.95 \\
        \hline
        LLaVa & CoT & 23.50 & 20.73 & 86.55 \\
    \bottomrule
    \end{tabular}
    \end{adjustbox}
    \caption{CoT Explanation quality evaluation with ground truth for AOKVQA dataset.}
    \label{tab:cot_gen_eval}
\end{table}
\textbf{Hallucination of Non-existent Objects}: In the case of identifying the room meant for rest, where the correct answer is a bathroom, LLaVA describes it as a bedroom containing a bed and a nightstand, exhibiting object hallucinations.

\textbf{Grounding Errors}: This type of mistake happens when the model incorrectly associates objects in a given context. For instance, when asked about the item on the bottom shelf near the TV, expected to be speakers, the generated rationale inaccurately identifies it as a remote control, demonstrating a clear grounding error.

\textbf{Inclusion of Irrelevant Details}: In examples like determining why people are waiting (where the correct answer is cross), the output includes details about a fire hydrant and emergency vehicles, which are not related to the original question.

\textbf{Language Priors}: The model may sometimes reference widely known subjects from the internet, which can lead to inaccuracies. For instance, in trying to identify the tennis player John McEnroe, the model unexpectedly mentions Roger Federer in its explanation. Federer, being one of the most famous tennis players in history, is a common topic online, suggesting that the model might be influenced by popular content. 

These examples highlight prevalent issues in generative models, including hallucinations, grounding inaccuracies, and the inclusion of irrelevant generic details, often influenced by language priors. These issues collectively hinder the effectiveness of the models in Chain-of-Thought tasks, suggesting a need for improved accuracy and relevance in visual comprehension and reasoning.

Furthermore, we conduct an automatic evaluation to assess the quality of generated CoT rationales in comparison to human-authored explanations. The results, presented in Table \ref{tab:cot_gen_eval}, demonstrate the performance of BF (XL) and BF(XXL) in zero-shot and few-shot ($n=5$) settings. We also tested the LLaVa model in the zero-shot scenario. Each model's performance is measured in terms of Rouge-1, Rouge-L, and BERTScore \cite{zhang2019bertscore}. Despite providing a numeric assessment, these automated metrics alone may not fully capture the quality of the generated rationales, indicating a need for further nuanced analysis. This limitation suggests that more in-depth analysis is required. Future research should focus on evaluating lengthy generative models' explanations in comparison to shorter, human-written ones.
\begin{table}[h]
\scriptsize
\begin{tabular}{p{0.2\textwidth}p{0.2\textwidth}p{0.5\textwidth}}
\textbf{Question \& Answer} &  \textbf{Ground truth} & \textbf{Generated rationale} \\
\midrule

What color is the floor? \textcolor{lightgray}{brown} 

\includegraphics[width=0.18\textwidth,height=0.1\textwidth,keepaspectratio]{}
& The floor is \textcolor{teal}{brown} wood flooring. 
& \vspace{-2mm} \begin{itemize}[leftmargin=*]
    \item \textbf{BLIP2}: The floor is made of wood. Wood is a light color. The answer: \textcolor{red}{light}.
    \item \textbf{LlaVa}: The floor in the image is \textcolor{red}{blue}.
\end{itemize} \\
\hline

\vspace{-1mm}
What item is on the bottom shelf near the TV? \textcolor{lightgray}{speakers} 

\includegraphics[width=0.18\textwidth,height=0.15\textwidth,keepaspectratio]{} 
& \vspace{-1mm} The shape of the device is in line with the object, and it is connected to a music player. 
& \vspace{-2mm} \begin{itemize}[leftmargin=*]
    \item \textbf{BLIP2}: A man and a woman are sitting on a couch watching tv. The man is holding a wii remote and the woman has a xbox controller. The bottom shelf near the TV has \textcolor{red}{snacks} on it.
    \item \textbf{LLaVa}: To determine the item on the bottom shelf near the TV, let's analyze the image step-by-step: 1. The image shows a man and a woman sitting on a couch in a living room. 2. There is a TV in the living room. 3. On the bottom shelf near the TV, there is an item. Based on these observations, it is likely that the item on the bottom shelf near the TV is a \textcolor{red}{remote control}. Remote controls are commonly placed near TVs for easy access and convenience.
\end{itemize} \\
\hline

\vspace{-1mm}
What room is pictured here for a rest? \textcolor{lightgray}{bathroom} 

\includegraphics[width=0.18\textwidth,height=0.15\textwidth,keepaspectratio]{} 
& \vspace{-1mm} The area is a bathroom since there is a soap dispenser. 
& \vspace{-2mm} \begin{itemize}[leftmargin=*]
    \item \textbf{BLIP2}: A \textcolor{teal}{bathroom} is a room for a rest. A bathroom has a sink, mirror, and toilet.
    \item \textbf{LLaVa}: The room pictured in the image appears to be a bedroom, as it contains a bed and a nightstand. The presence of these furniture items suggests that the room is designed for rest and relaxation. Additionally, the \textcolor{red}{bedroom} is decorated with a colorful mural on the wall, which adds an artistic touch to the space, making it visually appealing and inviting. Overall, the combination of the bed, nightstand, and colorful mural indicates that the room is intended for sleep and relaxation.
\end{itemize} \\
\hline

\vspace{-1mm}
What is the purpose of the item on the fridge handle? \textcolor{lightgray}{dry hands}

\includegraphics[width=0.18\textwidth,height=0.15\textwidth,keepaspectratio]{} 
& \vspace{-1mm} A small towel is wrapped along the bottom of the refrigerator's handle in this scene. given that towels are used to dry things and this towels proximity to the kitchen sink we can conclude this towel is usually used after \textcolor{teal}{washing hands}.
& \vspace{-2mm} \begin{itemize}[leftmargin=*]
    \item \textbf{BLIP2}:The item on the fridge handle is a \textcolor{red}{magnet}. Magnets are used to hold things in place.
    \item \textbf{LLaVa}:   To determine the purpose of the item on the fridge handle, let's analyze the image step-by-step: 1. The image shows a kitchen with a white refrigerator. 2. There is an item on the fridge handle. Based on these observations, it is likely that the item on the fridge handle is a \textcolor{red}{magnet}. Magnets are commonly used for various purposes, such as holding notes, reminders, or pictures on the fridge door.
\end{itemize} \\
\hline

\vspace{-1mm}
What is the name of a famous man associated with this sport? \textcolor{lightgray}{john mcenroe} 

\includegraphics[width=0.5\linewidth,keepaspectratio]{} 
& \vspace{-1mm} The two models in this ad are engaged in the game of tennis. \textcolor{teal}{mcenroe} is the only of the athletes listed associate with this sport.
& \vspace{-2mm} \begin{itemize}[leftmargin=*]
    \item \textbf{BLIP2}: \textcolor{red}{Federer} is a famous man associated with this sport. Federer was a professional tennis player.
    \item \textbf{LLaVa}: To determine the name of the famous man associated with this sport, we first need to identify the sport being played in the image. The image shows a woman playing tennis, so the sport in question is tennis.
    Next, we need to identify the famous man associated with tennis. One of the most well-known tennis players in history is Roger Federer, a Swiss professional tennis player who has won numerous Grand Slam titles and is considered one of the greatest tennis players of all time.
    Therefore, the name of the famous man associated with this sport is \textcolor{red}{Roger Federer.}
\end{itemize} \\
\hline

\vspace{-1mm}
What are they waiting to do when they stand next to the street? \textcolor{lightgray}{cross} 

\includegraphics[width=0.5\linewidth,keepaspectratio]{} 
& \vspace{-1mm} They are waiting to \textcolor{teal}{cross} the street.
& \vspace{-2mm} \begin{itemize}[leftmargin=*]
    \item \textbf{BLIP2}: They are waiting to \textcolor{teal}{cross} the street. They are standing next to a fire hydrant.
    \item \textbf{LLaVa}: Step-by-step reasoning:
1. The image shows a group of people standing next to a street.
2. There is a fire hydrant in the middle of the street.
3. The presence of the fire hydrant suggests that the people might be waiting for an emergency vehicle, such as a fire truck, to arrive at the scene.
4. Fire hydrants are essential for firefighters to access water for extinguishing fires.
5. Therefore, it is reasonable to assume that the people are waiting for an \textcolor{red}{emergency vehicle}, possibly a fire truck.
\end{itemize} \\
\hline

\vspace{-1mm}
What seems to be contained in the nook underneath the TV? \textcolor{lightgray}{fireplace} 

\includegraphics[width=0.5\linewidth,keepaspectratio]{} 
& \vspace{-1mm} There could be fire contained in the nook underneath the tv
& \vspace{-2mm} \begin{itemize}[leftmargin=*]
    \item \textbf{BLIP2}: The nook underneath the TV seems to be a \textcolor{red}{bookcase}. The bookcase contains a clock and a lamp.
    \item \textbf{LLaVa}: To determine what is contained in the nook underneath the TV, let's analyze the image step-by-step:
1. The image shows a bedroom with a bed, a TV, and a fireplace.
2. There is a nook underneath the TV.
3. The nook could potentially be used for storing items, such as books, DVDs, or other media.

Based on these observations, it is reasonable to conclude that the nook underneath the TV in the bedroom is likely used for storing items like \textcolor{red}{books, DVDs, or other items}.
\end{itemize} \\
\bottomrule
\end{tabular}
\caption{Comparison of ground truth explanations and CoT rationales for AOKVQA dataset}
\label{tab:cot_quality}
\end{table}

\subsection{Qualitative Samples: Model predictions across different prompt settings}
In this section, we provide an illustrative comparison of VQA answers generated by the BLIP2 model under different prompt settings. In Figure \ref{fig:example_images_aokvqa}, we present outputs from three different prompt settings. Each image is associated with a question, the ground truth response, a generated caption, as well as predictions from Standard-VQA, Caption-VQA, and CoT-VQA settings. 

The examples highlight both successful predictions and instances where the model deviates from the correct answer. We observe that Caption VQA tends to outperform Standard VQA in certain cases, leveraging the additional information present in captions to provide more accurate responses (see \textit{(c), (d)}). However, there are situations where relying solely on the caption may be insufficient, requiring the model to possess optical character recognition (OCR) capabilities that might be limited in BLIP2 models (see \textit{(e)}).

Furthermore, we note that CoT VQA exhibits some shortcomings, such as hallucination and generating lengthy rationales (see \textit{(a), (d)}). These factors contribute to a decrease in performance compared to the Standard VQA setting. Additionally, it's important to consider that the VQA accuracy metric may penalize answers that are correct but not an exact match to the ground truth answer (see \textit{(c)}).

Overall, this analysis emphasizes the impact of different prompt settings on the model's performance, highlighting the strengths and limitations of each approach in generating accurate VQA answers.

\begin{figure*}[h]

\newcommand{\imagebox}[2]{
    \begin{minipage}{0.46\linewidth}
        \begin{tcolorbox}[sharp corners,
                      colback=white,
                      boxrule=1.5pt,
                      left=1mm,  
                      right=1mm, 
                      top=1mm,   
                      bottom=1mm,
                      before skip=1mm, 
                      after skip=1mm]  
        \centering
        \includegraphics[width=0.7\linewidth,height=1.7in]{#1}
        \captionof*{figure}{#2}
    \end{tcolorbox}
    \end{minipage}
}
(\bf a)\imagebox{}{\scriptsize \textbf{Q}: \emph{Who might be helping the girl?} \textbf{A}: \emph{mother} \newline
          \textbf{Caption}: \emph{A girl holding a tray of food in a kitchen.} \newline
          \textbf{Standard-VQA}: \emph{\textcolor{teal}{her mother}} \newline
          \textbf{Caption-VQA}: \emph{\textcolor{teal}{her mother}} \newline
          \textbf{CoT-VQA}: \emph{The girl's mother might be helping the girl. \textcolor{brown}{The girl is in the kitchen with her mother. The mother is holding a tray of food.} The tray is full of cookies. Therefore, the final answer is \textcolor{teal}{a mother}.}}
\hfill
(\bf b)\imagebox{}{\scriptsize \textbf{Q}: \emph{What is the person being viewed through?}  \textbf{A}: \emph{window} \newline
          \textbf{Caption}: \emph{Two women sitting on a train looking at their phones.} \newline
          \textbf{Standard-VQA}: \emph{\textcolor{red}{windows}} \newline
          \textbf{Caption-VQA}: \emph{\textcolor{red}{windows}} \newline
          \textbf{CoT-VQA}: \emph{The person is being viewed through a \textcolor{teal}{window}. The window is on a train. The train is moving.}}
(\bf c)\imagebox{}{\scriptsize \newline \textbf{Q}: \emph{What type of pants is the man on the right wearing?} 
          \textbf{A}: \emph{jeans} \newline
          \textbf{Caption}: \emph{Two men standing in front of a window the man on the right is wearing jeans.} \\
          \textbf{Standard-VQA}: \emph{\textcolor{red}{black pants}} \newline
          \textbf{Caption-VQA}: \emph{\textcolor{teal}{jeans}} \newline
          \textbf{CoT-VQA}: \emph{The man on the right is wearing khaki pants. Khaki pants are a type of pants. Therefore, the final answer is \textcolor{red}{brown pants}.}}
(\bf d)\imagebox{}{\scriptsize \textbf{Q}: \emph{The girl is holding the pizza with what?} \textbf{A}: \emph{her mouth} \\
          \textbf{Caption}: \emph{A photo of a woman holding a piece of pizza in her mouth.} \\
          \textbf{Standard-VQA}: \emph{\textcolor{red}{hands}} \\
          \textbf{Caption-VQA}: \emph{\textcolor{teal}{her mouth }} \\
          \textbf{CoT-VQA}: \emph{The girl is holding the pizza with her hand. \textcolor{brown}{The pizza is on a paper plate. The paper plate is on top of a wooden table.} The table is in a restaurant. Therefore, the final answer is a \textcolor{red}{hand}.}}
(\bf e)\imagebox{}{\scriptsize \textbf{A}: \emph{ What type of cake are the donuts?}  
          \textbf{A}: \emph{carrot cake} \\
          \textbf{Caption}: \emph{A photo of three tiers of donuts on a counter.} \\
          \textbf{Standard-VQA}: \emph{\textcolor{red}{yeast}} \\
          \textbf{Caption-VQA}: \emph{\textcolor{red}{cake}} \\
          \textbf{CoT-VQA}: \emph{Donuts are a type of cake that are fried. The donuts in the picture are frosted. Frosting is made from sugar and water. The final answer: \textcolor{red}{cake}.}}    
\caption{Example images from the AOKVQA dataset with their respective Q\&A.}
\label{fig:example_images_aokvqa}
\end{figure*}

\end{document}